\documentclass{article}
\usepackage{ijcai23}

\usepackage{times}
\usepackage{soul}
\usepackage{url}
\usepackage[hidelinks]{hyperref}
\usepackage[utf8]{inputenc}
\usepackage[small]{caption}
\usepackage{graphicx}
\usepackage{amsmath}
\usepackage{amsthm}
\usepackage{booktabs}
\usepackage{algorithm}
\usepackage{algorithmic}
\usepackage{subcaption}
\usepackage[switch]{lineno}
\usepackage{xcolor}

\usepackage{array}
\newcolumntype{P}[1]{>{\centering\arraybackslash}p{#1}}

\title{MADS: Modulated Auto-Decoding SIREN for time series imputation}

\author{
Tom Bamford$^1$
\and
Elizabeth Fons$^1$\and
Yousef El-laham$^{1}$\and
Svitlana Vyetrenko$^1$
\affiliations
$^1$J.P. Morgan AI Research
\emails
tom.bamford@jpmchase.com
}

\begin{document}

\maketitle

\begin{abstract} 

Time series imputation remains a significant challenge across many fields due to the potentially significant variability in the type of data being modelled. 
Whilst traditional imputation methods often impose strong assumptions on the underlying data generation process, limiting their applicability, researchers have recently begun to investigate the potential of deep learning for this task, inspired by the strong performance shown by these models in both classification and regression problems across a range of applications. 
In this work we propose MADS, a novel auto-decoding framework for time series imputation, built upon implicit neural representations. Our method leverages the capabilities of SIRENs 
for high fidelity reconstruction of signals and irregular data, and combines it with a hypernetwork architecture which allows us to generalise by learning a prior over the space of time series. 
We evaluate our model on two real-world datasets, and show that it outperforms state-of-the-art methods for time series imputation. On the human activity dataset, it improves imputation performance by at least 40\%, while on the air quality dataset it is shown to be competitive across all metrics. When evaluated on synthetic data, our model results in the best average rank across different dataset configurations over all baselines.  
\end{abstract}

\section{Introduction}
\label{intro}

Time series imputation, and related tasks such as forecasting and generation, remain of significant interest in fields as diverse as finance, climate modelling and healthcare. Traditional approaches, such as averaging and regression, are typically over-simplistic and fail to adequately capture the underlying behaviour. The development of more modern methods, including iterative imputation and maximum likelihood routines, have increased the sophistication of the algorithms and improved performance, but the underlying assumptions often required by such approaches leads to implicit biases that can be detrimental in more complex cases (e.g. see \cite{Jarrett_HyperImpute} or \cite{Cao_BRITS}).

Implicit Neural Representations (INRs) have recently shown to have state-of-the-art performance in a variety of tasks, including shape representation, encoding of object appearance, part-level semantic segmentation and kernel representation (e.g. \cite{Mescheder_Occupancy_Nets}, \cite{Sitzmann_Scene_INR}, \cite{Kohli_Semantic_INR}, \cite{Romero_ckconv}). In particular, novel INR modifications such as periodic activations in SIREN (\cite{Sitzmann_INR_Periodic_Activation}) and positional encodings in NeRFs (\cite{Mildenhall_NeRF}) are able to overcome the spectral bias that traditional neural networks tend to suffer. Additionally, the grid-free learning approach compatible with INRs allow them to work well even for irregularly sampled or missing data.  

In this work, we propose a novel method for multivariate time series imputation via our Modulated Auto-Decoding SIREN (MADS). MADS utilises the capabilities of SIRENs for high fidelity reconstruction of signals and irregular data handling. Our SIREN parameterizations are combined with hypernetworks in order to learn a prior over the space of time series. Our contributions are summarised as follows:

\begin{itemize}
\item{We present Modulated Auto-Decoding SIREN (MADS) for time series imputation. MADS utilises a novel combination of amplitude modulation and SIREN weight prediction to allow notably robust performance across data regimes.} 
% \item{MADS utilises a novel combination of amplitude modulation and SIREN weight prediction to allow notably robust performance across data regimes}
\item{We propose a unique `fixed' formulation for amplitude modulation to aid generalisation across the dataset.}
\item{We evaluate our model on two real-world datasets, Air Quality and Human Activity Recognition and on multiple toy datasets that provide extensive coverage of different data regimes. Experimental results show that our model outperforms state-of-the-art models for time series imputation.}
\end{itemize}

\section{Related Work}
\label{related}

% \EF{ToDo: I'll finish the Related Work section}
\paragraph{Implicit Neural Representations} Implicit Neural Representations (INRs), also referred as neural fields, allow a continuous representation of multidimensional data by encoding a functional relationship between the input coordinates and their corresponding signal value. One of the main advantages of this representation is that the signal is encoded in a grid-free representation, providing an intrinsic non-linear interpolation of the data. One of the first applications of INRs was presented in DeepSDF for shape representation by a continuous volumetric field \cite{Park_DeepSDF}. 
DeepSDF is capable of representing an entire class of shapes through the use of an auto-decoder, and showed that a major advantage of the method is that inference can be performed with an arbitrary number of samples. This is particularly relevant to our case of time series imputation where the time series can be irregularly sampled and the number of samples can vary. More recently, INRs have gained popularity in visual computing (\cite{Mildenhall_NeRF}, \cite{Mescheder_Occupancy_Nets}) due to the key development of periodic activations \cite{Sitzmann_INR_Periodic_Activation} and positional encodings \cite{tancik2020fourfeat}, which allow them to learn high-frequency details within the data. 
Nevertheless, these methods can be computationally expensive or can have limited generalization capability as the complexity of the data increases (even with the use of hypernetworks). \cite{Mehta_Generalisable_INR} attempted to leverage periodic activations (SIREN), and their ability to reconstruct high frequency signals, whilst retaining generalisation capabilities, through the addition of a modulation network in the model architecture. The modulator is an additional MLP leveraged for generalisation, which consists of an identical internal structure to the SIREN (excepting the choice of activation function), such that each node output of the modulator can be matched up to a node in the SIREN, and element-wise multiplication carried out. 

Whilst there has been extensive uses of INRs in a wide variety of data sources such as video, images and audio, representation of 3D scene data, such as 3D geometry (\cite{Sitzmann_Scene_INR}, \cite{sitzmann2019metasdf}, \cite{Park_DeepSDF})
% ~\citep{park2019deepsdf, mescheder2019, sitzmann2019metasdf, sitzmann2019srns}
and object appearance (\cite{yariv2020multiview}, \cite{sztrajman2021}), few works have leveraged them for time series representation, with \cite{jeong2022} using INRs for anomaly detection and  \cite{woo2022deeptime} proposing INRs in combination with meta-learning for time series forecasting. Closest to our work, HyperTime \cite{Fons_HyperTime} implements a similar hypernetwork$+$SIREN architecture, but leverages it for time series generation. 

\paragraph{Time series imputation} Early time series imputation methods, which rely on basic statistical approaches, aim to leverage both the local continuity of the time dimension and the correlations among various channels. For example, the SimpleMean/SimpleMedian method \cite{Fung2006MethodsFT}, replaces missing values by taking the mean or median. The KNN (k-nearest neighbors) method \cite{Batista2002ASO}, uses cross-channel data to fill gaps with the help of k-nearest neighbours. In addition to the straightforward and standard interpolation techniques that utilise polynomial curve fitting, prevalent strategies focus on using established forecasting methods and drawing on the similarities between various time series to replace missing data points. For example, some approaches rely on k-nearest
neighbours (\cite{Troyanskaya2001MissingVE}, \cite{Beretta2016NearestNI}), the expectation-maximization
algorithm (\cite{Nelwamondo2007MissingDA}, \cite{Ghahramani1993SupervisedLF}) or linear predictors and state-space models (\cite{durbin2012time}, \cite{WalterO2013ImputationOI}).

Previous work has shown that deep learning models are able to capture the temporal dependency of time series, giving more accurate imputation than statistical methods. Common approaches use RNNs for sequence modelling (\cite{Cao_BRITS}, \cite{NEURIPS2018_96b9bff0}, \cite{Lipton16}, \cite{Yoon2017EstimatingMD}). 
Subsequent studies combined RNNs with other methods in order to improve imputation performance, such as GANs and self-training. In particular, the combination of RNNs with attention mechanisms have been successful for imputation and interpolation of time series. Whilst most time series imputation methods have focused on deterministic imputation, recently probabilistic methods have been developed, such as GP-VAE \cite{Fortuin2019GPVAEDP} and CSDI \cite{Tashiro_Diffusion_Model_Time_Series_Imputation}.

\section{Method}
\label{method}

\subsection{Problem Formulation}

In this work we aim to solve the problem of in-filling missing data via our proposed MADS model. We represent a general multivariate time series by a matrix of values $\mathbf{X} = {(\mathbf{x}_{0}, ..., \mathbf{x}_{N}})^{T} \in \mathcal{R}^{N,D}$ sampled at timesteps $\mathbf{T} = ({t_{0}, ..., t_{N}})$, where $D$ is the dimensionality of the series. Each column of this matrix therefore represents an individual time series corresponding to a feature of the original. Note that in general the timesteps are not regularly spaced. We can then define a corresponding mask matrix $\textbf{M} \in \{0\times1\}^{N,D}$, in which an element $M_{i,j}=0$ if the corresponding element in $\mathbf{X}$ is missing. In general there may be no pattern to these missing values, and all feature values at a given timestep could be missing.

\subsection{MADS} 

\begin{figure}
%\centering
\includegraphics[width=.9\linewidth]{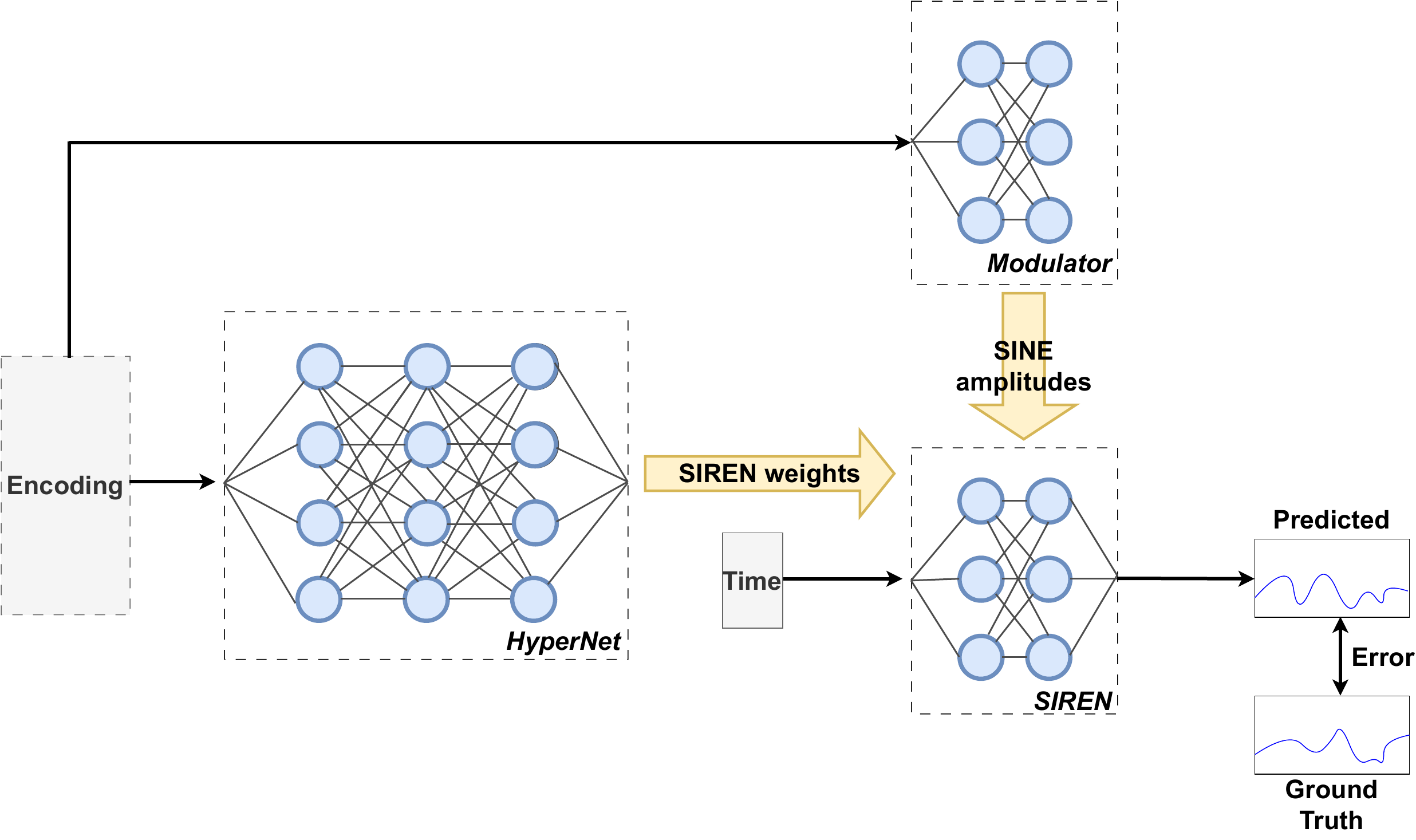}
\caption{MADS schematic.}
\label{MADS_Schematic}
\end{figure}

The architecture utilised by our model is shown in Fig. \ref{MADS_Schematic}. It consists of three networks: the foundational SIREN, a hypernet (\cite{Ha_Hypernetworks}), and a modulator. The hypernet takes in a latent code corresponding to a given time series, and outputs a set of network weights. These weights map to those inside the SIREN, and thus a distinct INR is instantiated for each individual time series from which imputation is carried out. Rather than utilise an encoding network to calculate the latent vectors, MADS follows the auto-decoding setup of DeepSDF (\cite{Park_DeepSDF}), which was found to be similarly capable without the additional network overhead. In this setup, the latent values are treated as variables during training (and so backpropagated), and then optimised again during inference for a given time series, when all other network variables are fixed. 

The third network is used for amplitude modulation, which is similar to that proposed in \cite{Mehta_Generalisable_INR}. Such a network was shown to improve generalisation performance significantly, relative to traditional auto-decoding or hypernetwork-based SIREN models. For a given sample, the modulator varies sine activation amplitudes within SIREN, allowing certain frequency modes to be nullified. Whilst the original model was applied to image-based data, and tasks such as up-sampling and image relighting, the applicability to time series and imputation in particular is clear.

In this work, a modulation network is leveraged alongside the hypernet. For a given time series this allows both SIREN weights and activation amplitudes to be varied, with the modulation network applying the weights through an element-wise mapping to the SIREN.

Defining $\textbf{W}$ and $\mathbf{b}$ as SIREN network weights and biases respectively, the output of the $i$'th hidden SIREN layer

\begin{equation}
h_{i} = \sin(\mathbf{W}_{i} h_{i-1}+\mathbf{b}_{i}) \ ,
\end{equation}

becomes

\begin{equation}
h_{i} = \alpha_{i} \odot \sin(\mathbf{W}_{i} h_{i-1}+\mathbf{b}_{i}) \ ,
\end{equation}

 such that the Hadamard product is applied between modulator hidden layer outputs $\alpha_{i}$ and the periodic activation function of SIREN. The modulation network is therefore constrained to having the same structure as the SIREN.

We investigate two formulations of MADS. The first, `base', formulation follows \cite{Mehta_Generalisable_INR}; the modulator takes in the same latent representation as that input into the hypernet, so that the modulated amplitudes vary with each individual time series. Since the modulator is introduced to limit overfitting, this can lead to amplitude variations for each time series that are too unconstrained. 
We therefore propose a more robust formulation that learns the frequency modes of the entire dataset rather than individual time series. In this setup, a distinct latent space is input into the modulator. As before, the latent variable values are learnt during training, but in this case no optimisation is carried out before inference, such that the latent values are shared across the full dataset. This is the `fixed' formulation.

To summarise, MADS constitutes the following features:

\begin{itemize}
\item{SIREN - functional representation of a given time series, taking an input timepoint and outputting corresponding amplitudes}
\item{Hypernet - takes in a latent vector representation of a time series and outputs a set of network weights used to instantiate the SIREN}
\item{Modulator - takes in a latent vector representation of a time series/dataset and modulates the sine activation amplitudes of SIREN through element-wise multiplication}
\item{Auto-decoding structure - treat latent variables as trainable parameters, and re-optimise during inference of a particular time series}
\end{itemize}

\section{Experiments}
\label{experiments}
% \label{methodology}

In this section we present the experiments used for model evaluation. We give a summary of the datasets used (section \ref{data}), briefly review the benchmarking models, and describe the metrics used for comparison (section \ref{benchmarks}). 
We also give the implementation details for MADS. 
We then go on to present our results and discuss their relevance (sections \ref{results_real} and \ref{results_toy}).

\subsection{Datasets}
\label{data}

\subsubsection{Real World Datasets}
We evaluate our model on two real world, multivariate datasets. The first one is the Human Activity dataset (HAR), which consists of three-dimensional spatial data collected from 5 people whilst carrying out a range of activities (sitting, walking, standing etc.). Four sensors were attached to each person - on their chest, belt and both ankles - giving 12 features in total. We follow the pre-processing scheme outlined in \cite{Rubanova_Latent_ODE} and \cite{Shukla_mTANS}, giving 6554 time series.

Following previous works \cite{Cao_BRITS}, \cite{Shukla_mTANS}, a specified fraction of known values are selected randomly for removal, acting as ground truths for imputation.
The missing data are selected randomly from each time series and the missing data fraction is user-specified; we evaluate two regimes of data with low/high missing data fractions, set here to 0.3 and 0.7 respectively.
We use the same randomly chosen missing timesteps for imputation across all features and time series. In addition, the dataset is split randomly into a train/test pair, with 20\% of the time series being designated to the test set. Note that each time series consists of 50 timesteps after pre-processing.

The second real world dataset we use is the Air Quality dataset, which consists of PM2.5 measurements taken from 36 measuring stations in Beijing, over a period of 12 months. The measurements were collected every hour. We follow the pre-processing steps of the data into a train-test split carried out in \cite{Yi_ST_MVL}. This is done for easy comparison with other works, which typically use the same split. 

In total, the dataset contains 158/80 samples for training/testing, with 36 features and the same number of timesteps. For both datasets, we run each experiment three times. 

\subsubsection{Toy Datasets}
\label{toy_data}

We construct the baseline sinusoidal toy time series (univariate) through the following expression:

\begin{equation}
y_{n} = e^{-\gamma (t_{n} + 1)} \sin(\Omega t_{n} + \phi) + \epsilon \\ , \ 
\end{equation}

where $\ \phi \sim \mathcal{N}(0,1) \ , \ \Omega \sim \omega \times \textrm{Beta}(2,2) \ , \ \epsilon \sim 0.2 \times \mathcal{N}(0,1)$, and $t_n \in \{-1, \dots, 1\}$. This is similar to the synthetic dataset used in \cite{Alaa_Fourier_Flows} when the decay factor $\gamma$ is set to zero. The exponential factor driven by $\gamma$ is included to model non-stationarity in the time series, whilst the beta distribution is scaled up by a constant multiplicative factor, $\omega$, to modify the frequency of modes within the dataset.

 To construct multivariate time series, the univariate equation is sampled multiple times to create a vector of independent time series, which can then be assigned as individual features. In this case, the amplitude is also taken from a Beta distribution, with all feature amplitudes re-scaled such that the maximum amplitude is 1. Typically, 200 timesteps are used to sample the time series, which is high enough to prevent aliasing in frequencies up to at least $\omega = 100$. 

 Note that the ground truth imputation values and train-test split follows that of the Human Activity dataset, i.e. missing values are randomly chosen and 20\% of the dataset is assigned to the test set. The total size of the dataset is set to 3000.

\subsection{Benchmarks and Evaluation}
\label{benchmarks}

We compare our model with classic imputation:
\begin{itemize}
    \item \textbf{Mean}: replace the missing values with the global mean.
    \item \textbf{Median} replace the missing values with the global median.
\end{itemize}
We also compare our model with existing state-of-the-art methods:
\begin{itemize}
    \item \textbf{CSDI} \cite{Tashiro_Diffusion_Model_Time_Series_Imputation}: utilizes score-based diffusion models conditioned on observed data, which is explicitly trained for imputation and is able to exploit correlations between observed values. 
    \item \textbf{BRITS} \cite{Cao_BRITS}: utilizes a bi-directional long short-term memory network  and can handle multiple correlated missing values in time series.
\end{itemize}

Additionally, we compare our method with three INR methods:
\begin{itemize}
    \item \textbf{Auto-SIREN} A single MLP with sine activations adapted from \cite{Park_DeepSDF}, with an additional input for the latent code $z$, concatenated with the coordinates $t$.
    \item \textbf{Mod-SIREN} A SIREN with fixed weights but conditionally predicted sine activation amplitudes via a modulation network and element-wise mapping between modulator weights and activation amplitudes. Here the auto-decoding variation is used which was proposed in the original work by \cite{Mehta_Generalisable_INR} .
    \item \textbf{HN+SIREN} A SIREN whose weights are conditionally predicted using a hypernetwork \cite{Ha_Hypernetworks}, similar to the hypertime architecture proposed for time series generation by \cite{Fons_HyperTime}. Instead of using a set-encoder, a hypernetwork takes the latent code $z$ as input, and predicts all the parameters of the SIREN.
\end{itemize}

\subsubsection{Metrics}

To assess model performance we use three metrics, which assess out-of-sample imputation accuracy. To evaluate, the model prior is set to the non-missing data of the test set, then evaluation is carried out on the missing points from the same dataset. The metrics used can be summarised as follows:

\begin{itemize}
\item{\textit{Imputation error (MSE):} MSE between the predicted (imputed) points in the test set compared to ground truth}
\item{\textit{Max imputation error (Max)}: measures the maximum error between missing data from the test set and those points evaluated using the trained model, then computes the average across all time series}
\item{\textit{Wasserstein2 (W2):} Euclidean distance in feature space between learnt distributions, pairing missing points in the ground truth and output set so as to minimise the total summed distance between pairs.}
\end{itemize}

\subsubsection{Implementation}

During training the model is optimised with respect to the following loss function,

\begin{equation}
\mathcal{L} = \mathcal{L}_{\textrm{mse}} + \mathcal{L}_{\textrm{latent}} + \mathcal{L}_{\textrm{weights}} \ ,
\label{eq:loss}
\end{equation}

which is consistent with that used in \cite{Fons_HyperTime}, \cite{Mehta_Generalisable_INR} and \cite{Park_DeepSDF}.
In e.q. \ref{eq:loss}, the first term is the standard MSE loss, defined via $\mathcal{L}_{\textrm{mse}} = \frac{1}{N} \sum^{N}_{i=1} (f_{i} - \hat{f}_{i})^{2}$, whilst the second and third terms are regularisation terms on the (hypernet) latent code and SIREN network weights respectively,

%\begin{equation}
\begin{align}
\mathcal{L}_{\textrm{latent}} &= \lambda_{Z} \frac{1}{N_{Z}} \sum\limits^{N_{Z}}_{j=1} z_{i}^{2} \\
\mathcal{L}_{\textrm{weights}} &= \lambda_{W} \frac{1}{N_{W}} \sum\limits^{N_{W}}_{k=1} W_{K}^{2} \ .
\end{align}
%\end{equation}

As in \cite{Fons_HyperTime}, the number of latent dimensions, $N_{Z}$, is set to 40. In addition, gradient clipping is used during training, which was found to improve performance. $\lambda_{Z}$ and $\lambda_{W}$ are set to 1e-01 and 1e-04 respectively. The latent space prior is set to the normal distribution $z \sim \mathcal{N}(0,\frac{1}{N_{W}})$, and following \cite{Park_DeepSDF} it was found that initialising the latent codes from a more compact distribution, $z \sim \mathcal{N}(0,0.01)$, during inference led to better generalisation performance. Optimisation is carried out using the Adam optimiser, with a learning rate of 5e-05 for model parameters, and 1e-03 for latent variables. Note that in the fixed formulation, only latent variables passing through the hypernet are regularised. The SIREN network consists of three fully-connected layers with hidden dimension of 60, whilst the hypernet is composed of a single hidden layers and a hidden feature size of 128. The modulator is constrained to the same structure as SIREN, with the activations replaced by ReLU functions. 
It should be noted that only a limited amount of hyper-parameter tuning was carried out for all models used in this work, with model setups primarily following that used in the original works.
% \EF{}

\subsection{Experimental Results on Real World datasets}
\label{results_real}

\begin{table*}[h!]
\label{sample-table}
\centering
% \begin{tabular}{l p{2cm}cccccr}
\resizebox{\textwidth}{!}{
% \begin{tabular}{lcccccccccp{5em}}
\begin{tabular}{lP{4em}P{4em}P{4em}P{4em}P{4em}P{4em}P{4em}P{4em}P{4em}}
\toprule
{} &  Mean &  Median &    BRITS &    CSDI &  Auto-SIREN &  Mod-SIREN &  HN+ SIREN &     MADS-Base (Ours) &  MADS-Fixed (Ours)\\
 \midrule
\textit{Air Quality} & & & & & & \\ 
 \midrule
MSE     &     9.1e-2 &     9.1e-2 & 2.1e-1  & \bf 6.7e-2 &     9.1e-2 &    7.2e-2 &         8.1e-2 & 7.6e-2 &     7.3e-2 \\
        &     (0.0e+0) &     (0.0e+0) & (5.8e-3)  & (4.4e-3) &     (5.3e-3) &    (2.9e-3) &  (1.1e-2) & (2.3e-3) &   (1.1e-2) \\
Max     &     \bf 4.7e-1 &     4.8e-1 & 8.7e-1  & 8.3e-1 &     5.2e-1 &    4.8e-1 &         5.2e-1 & 5.1e-1 &     4.9e-1 \\
        &     (0.0e+0) &     (0.0e+0) & (1.0e-2)  & (5.8e-3) &     (2.9e-2) &    (1.0e-2) &  (1.0e-2) & (3.5e-2) &  (2.0e-2) \\
W2      &     2.6e-1 &     2.6e-1 & 3.7e-1  & 1.2e+0 &     2.1e-1 &    2.4e-1 &         2.3e-1 & 2.3e-1 &     \bf 2.0e-1 \\
        &     (0.0e+0) &     (0.0e+0) & (5.8e-3)  & (0.0e+0) &     (2.1e-2) &  (2.5e-2) &   (1.0e-2) & (3.8e-2) &  (1.5e-2) \\
\midrule
\textit{HAR ($30\%$)} & & & & & & \\ 
 \midrule
MSE     &     2.8e-2 &     2.9e-2 & 3.1e-1     & 9.3e-3     &     6.4e-3        &    6.2e-3     &   1.1e-2      & 5.4e-3    &     \bf 5.3e-3 \\
        &     (5.8e-4) &     (0.0e+0) & (0.0e+0)   & (1.7e-3)   &     (8.2e-4)      &    (1.5e-3)   &   (5.9e-3)    & (2.6e-4)  &     (5.1e-4) \\
Max     &     3.3e-1 &     3.3e-1 & 9.5e-1     & 2.8e-1     &     2.1e-1        &    2.1e-1     &   2.9e-1      & 2.0e-1    &     \bf 2.0e-1 \\
        &     (5.8e-3) &     (5.8e-3) & (5.8e-3)   & (2.1e-2)   &     (1.0e-2)      &    (1.2e-2)   &   (9.5e-2)    & (5.8e-3)  &     (1.0e-2) \\
W2      &     7.8e-2 &     7.8e-2 & 2.9e-1     & 1.1e-1     &     1.0e-1        &    9.4e-2     &   1.3e-1      & 8.5e-2    &     \bf 7.8e-2 \\
        &     (5.8e-4) &     (5.8e-4) & (5.8e-3)   & (1.6e-2)   &     (1.3e-2)      &    (1.7e-2)   &   (4.2e-2)    & (1.7e-3)  &     (4.4e-3) \\
 \midrule
\textit{HAR ($70\%$)} & & & & & & \\ 
 \midrule
MSE     &     2.8e-2 &     2.9e-2 & 3.1e-1     & 2.1e-2     & \bf 6.0e-3    &    1.3e-2     &   1.5e-2      & 6.5e-3    &     \bf 6.0e-3 \\
        &     (0.0e+0) &     (0.0e+0) & (5.8e-3)   & (6.7e-3)   &     (5.7e-4)  &    (1.2e-2)   &   (6.7e-3)    & (5.6e-4)  &     (2.9e-4) \\
Max     &     3.6e-1 &     3.6e-1 & 9.9e-1    & 4.5e-1     & \bf 2.5e-1    &    2.9e-1     &   3.9e-1      & 2.7e-1    &     2.6e-1 \\
        &     (0.0e+0) &     (5.8e-3) & (0.0e+0)  & (6.0e-2)   &     (5.8e-3)  &    (5.8e-2)   &   (7.0e-2)    & (5.8e-3)  &     (5.8e-3) \\
W2      &     1.8e-1 &     1.8e-1 & 6.7e-1     & 1.7e-1     &     9.9e-2    &    1.4e-1     &   1.6e-1      & 8.6e-2    &    \bf 8.3e-2 \\
        &     (5.8e-3) &     (0.0e+0) & (0.0e+0)  & (3.6e-2)   &     (1.2e-2)  &    (9.4e-2)   &   (4.4e-2)    & (4.5e-3)  &     (2.1e-3) \\
\bottomrule
\end{tabular}
}
\caption{Comparison of MADS with alternative deep learning benchmarks for the imputation task on unseen data. Results are averaged over three runs, with the standard deviation also shown for the imputed MSE.}
\label{tab:real_datasets_result}
\end{table*}

Table \ref{tab:real_datasets_result} shows the imputation results evaluated with the real world datasets. MADS outperforms the baselines on both configurations of HAR dataset and shows competitive performance with CSDI on the Air Quality dataset. Additionally, it outperforms all baselines and all datasets on W2 metric. 
Both formulations of MADS perform competitively across the board, with the fixed variant notably out-performing all benchmarks in the majority of metrics. In addition, for both air quality metrics in which it was beaten the optimal value is within the uncertainty range of the fixed formulation. The base MADS formulation also performs well relative to the benchmarks, but for these datasets is unable to out-perform its sibling on any particular metric. 
The two datasets correspond to different data regimes: HAR showing less variation with fewer features and a changing missing ratio, and air quality a complex dataset with correlated missing values. MADS therefore shows robust performance in both settings, particularly given the limited amount of hyper-parameter tuning carried out. The model is also typically quicker to train than both BRITS and CSDI, achieving convergence in only a few minutes. Note that relative to the other models modulated SIREN shows significant instability during training.

\subsection{Experimental Results on Toy Dataset}

\label{results_toy}
\begin{table}[t]
\centering
\begin{tabular}{l | cccr}
\toprule
 Dataset & \multicolumn4r{Data regime} \\
\midrule
 & Freq. & Dim. & Noise & Length \\
 \hline
 B-SIN & L & L & $\times$ & $\times$ \\
 M-SIN & N & L & $\times$ & $\times$ \\
 F-SIN & H & L & $\times$ & $\times$ \\
 D-SIN & L & H & $\times$ & $\times$ \\
 MD-SIN & N & H & $\times$ & $\times$ \\
 FD-SIN & H & H & $\times$ & $\times$ \\
 N-SIN & L & L & $\surd$ & $\times$ \\
 L-SIN & H & L & $\times$ & $\surd$ \\
\bottomrule
\end{tabular}
\caption{Toy datasets used to investigate different data regimes.}
\label{tab:toy_datasets}
\end{table}

In the second set of experiments, the synthetic dataset introduced in section \ref{toy_data} is utilised to systematically control the data regime and thus compare model performance across scenarios. The model parameters $\omega$ and $\gamma$ are varied along with dimensionality and the addition of noise to assess relative performance. The following datasets, summarised in Table. \ref{tab:toy_datasets}, are constructed:

\begin{table*}[h!]
\label{sample-table}
\centering
\begin{tabular}{l | ccccccr}
\toprule
 Dataset & BRITS & CSDI & Auto-SIREN & Mod-SIREN & HN+ SIREN & MADS & MADS (f) \\
 \midrule
 B-SIN & 3.5e-01 & \textbf{3.2e-04} & 1.2e-02 & 2.2e-03 & 8.0e-03 & 3.1e-03 & 6.0e-03 \\
 M-SIN & 3.6e-01 & 9.7e-01 & 2.4e-01 & 3.5e-01 & 1.9e-01 & \textbf{9.8e-02} & 1.0e-01 \\
 F-SIN & \textbf{3.3e-01} & 9.8e-01 & 5.0e-01 & 5.0e-01 & 4.5e-01 & 3.8e-01 & 3.7e-01 \\
 D-SIN & 3.7e-01 & \textbf{1.4e-04} & 2.1e-01 & 2.2e-01 & 1.3e-01 & 9.1e-02 & 8.4e-02\\
 MD-SIN & \textbf{1.5e-01} & 9.7e-01 & 4.6e-01 & 4.3e-01 & 4.3e-01 & 3.7e-01 & 3.6e-01 \\
 FD-SIN & \textbf{4.1e-01} & 9.8e-01 & 5.1e-01 & 5.1e-01 & 5.5e-01 & 5.2e-01 & 5.3e-01 \\
 N-SIN & 1.4e-01 & 3.4e-02 & \textbf{2.5e-02} & \textbf{2.5e-02} & 4.0e-02 & 3.4e-02 & 3.5e-02 \\
 L-SIN & \textbf{7.4e-02} & 2.7e-01 & 1.3e-01 & 1.4e-01 & 1.1e-01 & 8.6e-02 & 8.5e-02 \\
 \midrule
 Avg rank & 3.9 & 5.0 & 4.3 & 3.9 & 4.5 & \textbf{2.9} & 3.0 \\
\bottomrule
\end{tabular}
\caption{Imputation error (MSE) for the deep learning models with the toy datasets.}
\label{tab:toy_datasets_result}
\end{table*}

\begin{itemize}
\item{\textbf{B-SIN}: baseline dataset with low frequency ($\omega=5$) and dimensionality (2)}
\item{\textbf{M-SIN}: mid frequency dataset ($\omega=30$) with low dimensionality (2)}
\item{\textbf{F-SIN}: high frequency dataset ($\omega=100$) with low dimensionality (2)}
\item{\textbf{D-SIN}: high dimensionality dataset (10) with low frequency ($\omega=5$)}
\item{\textbf{MD-SIN}: mid frequency ($\omega=30$), high dimensionality (10)}
\item{\textbf{FD-SIN}: high frequency ($\omega=100$), high dimensionality (10)}
\item{\textbf{N-SIN}: noisy dataset with low frequency ($\omega=5$) and  low dimensionality (2)}
\item{\textbf{L-SIN}: long/time-dependent dataset ($\gamma=1$) with high frequency ($\omega=100$) and low dimensionality (2)}
\end{itemize}

All datasets have no time dependence and zero noise unless explicitly stated. As explained in section \ref{toy_data}, missing data is generated using random sampling, following the same approach as for the Human Activity datasetm with the missingness fraction set to 0.3.

Table \ref{tab:toy_datasets_result} shows model performance results for the different configurations of the toy datasets, via the imputation MSE. As with the real datasets, MADS is found to perform competitively across all cases, and indeed both variants achieves the highest averaged rankings across all datasets. In these experiments the base MADS formulation performs slightly better than the fixed version, although the fixed variant still appears to out-perform its sibling for a higher frequency or high dimensional dataset. 

MADS is a notably more consistent performer across data regimes relative to the other approaches. Unlike these alternative models, the underlying assumptions on the dataset made by the INR-based approach is limited. This contrasts with CSDI, which outperforms all competitors for the simple baseline and high dimensionality cases, but struggles with higher frequency signals. It has been established for this model (see e.g. \cite{Chen_Noise_Scheduling_Diffusion_Models}, \cite{Hoogeboom_Simple_Diffusion}) that the assumptions made on the noise scheduling setup of the training process can require significant amounts of tuning for different datasets, which potentially explains the variability in performance seen here. Similarly BRITS shows very different performance capabilities across datasets, seemingly excelling for the high frequency and long/non-stationary time series, but performing very poorly for the other cases. 
This is likely due to assumptions built into the model. For example, the assumption of causality implied by a recurrent model may ensure good performance in the non-stationary case, but the delayed backpropagation utilised to allow learning from imputed values during this process is likely to limit reconstruction capabilities and thus high performance in simpler cases.

\section{Conclusions}

In this work, we proposed MADS, a novel method that uses INRs within a modulated hypernetwork architecture for time series imputation. The proposed model is compared to state-of-the-art benchmarks within the deep learning literature, across both real-world and toy multivariate datasets. We found the model to be robust across different data regimes and to outperform alternative approaches in the majority of metrics on the real-world datasets. In addition, across toy datasets the model is shown to  have the highest average ranking relative to all baseline models. 

\paragraph{Disclaimer}
This paper was prepared for informational purposes by
the Artificial Intelligence Research group of JPMorgan Chase \& Co\. and its affiliates (``JP Morgan''),
and is not a product of the Research Department of JP Morgan.
JP Morgan makes no representation and warranty whatsoever and disclaims all liability,
for the completeness, accuracy or reliability of the information contained herein.
This document is not intended as investment research or investment advice, or a recommendation,
offer or solicitation for the purchase or sale of any security, financial instrument, financial product or service,
or to be used in any way for evaluating the merits of participating in any transaction,
and shall not constitute a solicitation under any jurisdiction or to any person,
if such solicitation under such jurisdiction or to such person would be unlawful.

\bibliographystyle{named}
\bibliography{ijcai23}

\begin{thebibliography}{}

\bibitem[\protect\citeauthoryear{Alaa \bgroup \em et al.\egroup
  }{2021}]{Alaa_Fourier_Flows}
Ahmed Alaa, Alex~James Chan, and Mihaela van~der Schaar.
\newblock Generative time-series modeling with fourier flows.
\newblock In {\em International Conference on Learning Representations}, 2021.

\bibitem[\protect\citeauthoryear{Batista and Monard}{2002}]{Batista2002ASO}
Gustavo E. A. P.~A. Batista and Maria~Carolina Monard.
\newblock A study of k-nearest neighbour as an imputation method.
\newblock In {\em International Conference on Health Information Science},
  2002.

\bibitem[\protect\citeauthoryear{Beretta and
  Santaniello}{2016}]{Beretta2016NearestNI}
Lorenzo Beretta and Alessandro Santaniello.
\newblock Nearest neighbor imputation algorithms: a critical evaluation.
\newblock {\em BMC Medical Informatics and Decision Making}, 16, 2016.

\bibitem[\protect\citeauthoryear{Cao \bgroup \em et al.\egroup
  }{2018}]{Cao_BRITS}
Wei Cao, Dong Wang, Jian Li, Hao Zhou, Lei Li, and Yitan Li.
\newblock Brits: Bidirectional recurrent imputation for time series.
\newblock In S.~Bengio, H.~Wallach, H.~Larochelle, K.~Grauman, N.~Cesa-Bianchi,
  and R.~Garnett, editors, {\em Advances in Neural Information Processing
  Systems}, volume~31. Curran Associates, Inc., 2018.

\bibitem[\protect\citeauthoryear{Chen}{2023}]{Chen_Noise_Scheduling_Diffusion_Models}
Ting Chen.
\newblock On the importance of noise scheduling for diffusion models, 2023.

\bibitem[\protect\citeauthoryear{Durbin and Koopman}{2012}]{durbin2012time}
T.J. Durbin and S.J. Koopman.
\newblock {\em Time Series Analysis by State Space Methods: Second Edition}.
\newblock Oxford Statistical Science Series. OUP Oxford, 2012.

\bibitem[\protect\citeauthoryear{Fons \bgroup \em et al.\egroup
  }{2022}]{Fons_HyperTime}
Elizabeth Fons, Alejandro Sztrajman, Yousef El-laham, Alexandros Iosifidis, and
  Svitlana Vyetrenko.
\newblock Hypertime: Implicit neural representation for time series, 2022.

\bibitem[\protect\citeauthoryear{Fortuin \bgroup \em et al.\egroup
  }{2019}]{Fortuin2019GPVAEDP}
Vincent Fortuin, Dmitry Baranchuk, Gunnar R{\"a}tsch, and Stephan Mandt.
\newblock Gp-vae: Deep probabilistic time series imputation.
\newblock In {\em International Conference on Artificial Intelligence and
  Statistics}, 2019.

\bibitem[\protect\citeauthoryear{Fung}{2006}]{Fung2006MethodsFT}
David~S. Fung.
\newblock Methods for the estimation of missing values in time series.
\newblock 2006.

\bibitem[\protect\citeauthoryear{Ghahramani and
  Jordan}{1993}]{Ghahramani1993SupervisedLF}
Zoubin Ghahramani and Michael~I. Jordan.
\newblock Supervised learning from incomplete data via an em approach.
\newblock In {\em NIPS}, 1993.

\bibitem[\protect\citeauthoryear{Ha \bgroup \em et al.\egroup
  }{2016}]{Ha_Hypernetworks}
David Ha, Andrew Dai, and Quoc~V. Le.
\newblock Hypernetworks, 2016.

\bibitem[\protect\citeauthoryear{Hoogeboom \bgroup \em et al.\egroup
  }{2023}]{Hoogeboom_Simple_Diffusion}
Emiel Hoogeboom, Jonathan Heek, and Tim Salimans.
\newblock simple diffusion: End-to-end diffusion for high resolution images,
  2023.

\bibitem[\protect\citeauthoryear{Jarrett \bgroup \em et al.\egroup
  }{2022}]{Jarrett_HyperImpute}
Daniel Jarrett, Bogdan Cebere, Tennison Liu, Alicia Curth, and Mihaela van~der
  Schaar.
\newblock Hyperimpute: Generalized iterative imputation with automatic model
  selection.
\newblock 2022.

\bibitem[\protect\citeauthoryear{Jeong and Shin}{2022}]{jeong2022}
Kyeong{-}Joong Jeong and Yong{-}Min Shin.
\newblock Time-series anomaly detection with implicit neural representation.
\newblock {\em CoRR}, abs/2201.11950, 2022.

\bibitem[\protect\citeauthoryear{Kohli \bgroup \em et al.\egroup
  }{2020}]{Kohli_Semantic_INR}
Amit Kohli, Vincent Sitzmann, and Gordon Wetzstein.
\newblock Inferring semantic information with 3d neural scene representations.
\newblock {\em ArXiv}, abs/2003.12673, 2020.

\bibitem[\protect\citeauthoryear{Lipton \bgroup \em et al.\egroup
  }{2016}]{Lipton16}
Zachary~C Lipton, David Kale, and Randall Wetzel.
\newblock Directly modeling missing data in sequences with rnns: Improved
  classification of clinical time series.
\newblock In Finale Doshi-Velez, Jim Fackler, David Kale, Byron Wallace, and
  Jenna Wiens, editors, {\em Proceedings of the 1st Machine Learning for
  Healthcare Conference}, volume~56 of {\em Proceedings of Machine Learning
  Research}, pages 253--270, Northeastern University, Boston, MA, USA, 18--19
  Aug 2016. PMLR.

\bibitem[\protect\citeauthoryear{Luo \bgroup \em et al.\egroup
  }{2018}]{NEURIPS2018_96b9bff0}
Yonghong Luo, Xiangrui Cai, Ying ZHANG, Jun Xu, and Yuan xiaojie.
\newblock Multivariate time series imputation with generative adversarial
  networks.
\newblock In S.~Bengio, H.~Wallach, H.~Larochelle, K.~Grauman, N.~Cesa-Bianchi,
  and R.~Garnett, editors, {\em Advances in Neural Information Processing
  Systems}, volume~31. Curran Associates, Inc., 2018.

\bibitem[\protect\citeauthoryear{Mehta \bgroup \em et al.\egroup
  }{2021}]{Mehta_Generalisable_INR}
Ishit Mehta, Micha\"el Gharbi, Connelly Barnes, Eli Shechtman, Ravi
  Ramamoorthi, and Manmohan Chandraker.
\newblock Modulated periodic activations for generalizable local functional
  representations.
\newblock In {\em Proceedings of the IEEE/CVF International Conference on
  Computer Vision (ICCV)}, pages 14214--14223, October 2021.

\bibitem[\protect\citeauthoryear{Mescheder \bgroup \em et al.\egroup
  }{2019}]{Mescheder_Occupancy_Nets}
Lars Mescheder, Michael Oechsle, Michael Niemeyer, Sebastian Nowozin, and
  Andreas Geiger.
\newblock Occupancy networks: Learning 3d reconstruction in function space.
\newblock In {\em Proceedings IEEE Conf. on Computer Vision and Pattern
  Recognition (CVPR)}, 2019.

\bibitem[\protect\citeauthoryear{Mildenhall \bgroup \em et al.\egroup
  }{2020}]{Mildenhall_NeRF}
Ben Mildenhall, Pratul~P. Srinivasan, Matthew Tancik, Jonathan~T. Barron, Ravi
  Ramamoorthi, and Ren Ng.
\newblock Nerf: Representing scenes as neural radiance fields for view
  synthesis.
\newblock In {\em ECCV}, 2020.

\bibitem[\protect\citeauthoryear{Nelwamondo \bgroup \em et al.\egroup
  }{2007}]{Nelwamondo2007MissingDA}
Fulufhelo~Vincent Nelwamondo, Shakir Mohamed, and T.~Marwala.
\newblock Missing data: A comparison of neural network and expectation
  maximization techniques.
\newblock {\em Current Science}, 93:1514--1521, 2007.

\bibitem[\protect\citeauthoryear{Park \bgroup \em et al.\egroup
  }{2019}]{Park_DeepSDF}
Jeong~Joon Park, Peter Florence, Julian Straub, Richard Newcombe, and Steven
  Lovegrove.
\newblock Deepsdf: Learning continuous signed distance functions for shape
  representation.
\newblock In {\em The IEEE Conference on Computer Vision and Pattern
  Recognition (CVPR)}, June 2019.

\bibitem[\protect\citeauthoryear{Romero \bgroup \em et al.\egroup
  }{2022}]{Romero_ckconv}
David~W. Romero, Anna Kuzina, Erik~J Bekkers, Jakub~Mikolaj Tomczak, and Mark
  Hoogendoorn.
\newblock {CKC}onv: Continuous kernel convolution for sequential data.
\newblock In {\em International Conference on Learning Representations}, 2022.

\bibitem[\protect\citeauthoryear{Rubanova \bgroup \em et al.\egroup
  }{2019}]{Rubanova_Latent_ODE}
Yulia Rubanova, Ricky T.~Q. Chen, and David~K Duvenaud.
\newblock Latent ordinary differential equations for irregularly-sampled time
  series.
\newblock In H.~Wallach, H.~Larochelle, A.~Beygelzimer, F.~d\textquotesingle
  Alch\'{e}-Buc, E.~Fox, and R.~Garnett, editors, {\em Advances in Neural
  Information Processing Systems}, volume~32. Curran Associates, Inc., 2019.

\bibitem[\protect\citeauthoryear{Shukla and Marlin}{2021}]{Shukla_mTANS}
Satya~Narayan Shukla and Benjamin Marlin.
\newblock Multi-time attention networks for irregularly sampled time series.
\newblock In {\em International Conference on Learning Representations}, 2021.

\bibitem[\protect\citeauthoryear{Sitzmann \bgroup \em et al.\egroup
  }{2019}]{Sitzmann_Scene_INR}
Vincent Sitzmann, Michael Zollh\"{o}fer, and Gordon Wetzstein.
\newblock {\em Scene Representation Networks: Continuous 3D-Structure-Aware
  Neural Scene Representations}.
\newblock Curran Associates Inc., Red Hook, NY, USA, 2019.

\bibitem[\protect\citeauthoryear{Sitzmann \bgroup \em et al.\egroup
  }{2020a}]{sitzmann2019metasdf}
Vincent Sitzmann, Eric~R. Chan, Richard Tucker, Noah Snavely, and Gordon
  Wetzstein.
\newblock Metasdf: Meta-learning signed distance functions.
\newblock In {\em Proc. NeurIPS}, 2020.

\bibitem[\protect\citeauthoryear{Sitzmann \bgroup \em et al.\egroup
  }{2020b}]{Sitzmann_INR_Periodic_Activation}
Vincent Sitzmann, Julien N.~P. Martel, Alexander~W. Bergman, David~B. Lindell,
  and Gordon Wetzstein.
\newblock Implicit neural representations with periodic activation functions.
\newblock {\em CoRR}, abs/2006.09661, 2020.

\bibitem[\protect\citeauthoryear{Sztrajman \bgroup \em et al.\egroup
  }{2021}]{sztrajman2021}
Alejandro Sztrajman, Gilles Rainer, Tobias Ritschel, and Tim Weyrich.
\newblock Neural brdf representation and importance sampling.
\newblock {\em Computer Graphics Forum}, 40(6):332--346, 2021.

\bibitem[\protect\citeauthoryear{Tancik \bgroup \em et al.\egroup
  }{2020}]{tancik2020fourfeat}
Matthew Tancik, Pratul~P. Srinivasan, Ben Mildenhall, Sara Fridovich-Keil,
  Nithin Raghavan, Utkarsh Singhal, Ravi Ramamoorthi, Jonathan~T. Barron, and
  Ren Ng.
\newblock Fourier features let networks learn high frequency functions in low
  dimensional domains.
\newblock {\em NeurIPS}, 2020.

\bibitem[\protect\citeauthoryear{Tashiro \bgroup \em et al.\egroup
  }{2021}]{Tashiro_Diffusion_Model_Time_Series_Imputation}
Yusuke Tashiro, Jiaming Song, Yang Song, and Stefano Ermon.
\newblock Csdi: Conditional score-based diffusion models for probabilistic time
  series imputation.
\newblock In M.~Ranzato, A.~Beygelzimer, Y.~Dauphin, P.S. Liang, and J.~Wortman
  Vaughan, editors, {\em Advances in Neural Information Processing Systems},
  volume~34, pages 24804--24816. Curran Associates, Inc., 2021.

\bibitem[\protect\citeauthoryear{Troyanskaya \bgroup \em et al.\egroup
  }{2001}]{Troyanskaya2001MissingVE}
Olga~G. Troyanskaya, Michael~N. Cantor, Gavin Sherlock, Patrick~O. Brown,
  Trevor~J. Hastie, Robert Tibshirani, David Botstein, and Russ~B. Altman.
\newblock Missing value estimation methods for dna microarrays.
\newblock {\em Bioinformatics}, 17 6:520--5, 2001.

\bibitem[\protect\citeauthoryear{Walter.O \bgroup \em et al.\egroup
  }{2013}]{WalterO2013ImputationOI}
Yodah Walter.O, John Kihoro, K.H.O Athiany, and W~KibunjaH.
\newblock Imputation of incomplete non- stationary seasonal time series data.
\newblock {\em Mathematical theory and modeling}, 3:142--154, 2013.

\bibitem[\protect\citeauthoryear{Woo \bgroup \em et al.\egroup
  }{2022}]{woo2022deeptime}
Gerald Woo, Chenghao Liu, Doyen Sahoo, Akshat Kumar, and Steven C.~H. Hoi.
\newblock Deeptime: Deep time-index meta-learning for non-stationary
  time-series forecasting.
\newblock 2022.

\bibitem[\protect\citeauthoryear{Yariv \bgroup \em et al.\egroup
  }{2020}]{yariv2020multiview}
Lior Yariv, Yoni Kasten, Dror Moran, Meirav Galun, Matan Atzmon, Basri Ronen,
  and Yaron Lipman.
\newblock Multiview neural surface reconstruction by disentangling geometry and
  appearance.
\newblock {\em Advances in Neural Information Processing Systems}, 33, 2020.

\bibitem[\protect\citeauthoryear{Yi \bgroup \em et al.\egroup
  }{2016}]{Yi_ST_MVL}
Xiuwen Yi, Yu~Zheng, Junbo Zhang, and Tianrui Li.
\newblock St-mvl: Filling missing values in geo-sensory time series data.
\newblock In {\em International Joint Conference on Artificial Intelligence},
  2016.

\bibitem[\protect\citeauthoryear{Yoon \bgroup \em et al.\egroup
  }{2017}]{Yoon2017EstimatingMD}
Jinsung Yoon, William~R. Zame, and Mihaela van~der Schaar.
\newblock Estimating missing data in temporal data streams using
  multi-directional recurrent neural networks.
\newblock {\em IEEE Transactions on Biomedical Engineering}, 66:1477--1490,
  2017.

\end{thebibliography}

\end{document}